\documentclass[letterpaper]{article} 
\usepackage{aaai25}  
\usepackage{times}  
\usepackage{helvet}  
\usepackage{courier}  
\usepackage[hyphens]{url}  
\usepackage{graphicx} 
\usepackage{natbib}  
\usepackage{caption} 
\usepackage{amsfonts}
\usepackage{amsmath}
\usepackage{booktabs}
\usepackage{newtxtext,newtxmath}
\usepackage{tcolorbox}
\usepackage{xcolor}
\definecolor{lightgray}{rgb}{0.95, 0.95, 0.95}
\usepackage{newfloat}
\usepackage{listings}
\DeclareCaptionStyle{ruled}{labelfont=normalfont,labelsep=colon,strut=off} 
\nocopyright 

\title{ReXTrust: A Model for Fine-Grained Hallucination Detection in AI-Generated Radiology Reports}
\author {
    Romain Hardy\textsuperscript{\rm 1},
    Sung Eun Kim\textsuperscript{\rm 1},
    Du Hyun Ro\textsuperscript{\rm 2},
    Pranav Rajpurkar\textsuperscript{\rm 1}
}
\affiliations {
    \textsuperscript{\rm 1}Department of Biomedical Informatics, Harvard Medical School\\
    \textsuperscript{\rm 2}Department of Orthopedic Surgery, Seoul National University Hospital\\
    romain\_hardy@fas.harvard.edu, sungeun\_kim2@hms.harvard.edu, \\
    franciu7@snu.ac.kr,
    pranav\_rajpurkar@hms.harvard.edu
}

\begin{document}

\maketitle

\begin{abstract}
The increasing adoption of AI-generated radiology reports necessitates robust methods for detecting hallucinations—false or unfounded statements that could impact patient care. We present ReXTrust, a novel framework for fine-grained hallucination detection in AI-generated radiology reports. Our approach leverages sequences of hidden states from large vision-language models to produce finding-level hallucination risk scores. We evaluate ReXTrust on a subset of the MIMIC-CXR dataset and demonstrate superior performance compared to existing approaches, achieving an AUROC of 0.8751 across all findings and 0.8963 on clinically significant findings. Our results show that white-box approaches leveraging model hidden states can provide reliable hallucination detection for medical AI systems, potentially improving the safety and reliability of automated radiology reporting.
\end{abstract}

\section{Introduction}
The automated generation of radiology reports using large vision-language models (LVLMs) has recently shown promising results, offering the potential to improve workflow efficiency and standardization in clinical settings \citep{bannur2024maira, tanida2023interactive, hamamci2024ct2rep, liu2024bootstrapping, gu2024complex, chen2024chexagent, zhou2024generalist}. However, these models sometimes generate hallucinations—statements that are false, unfounded, or inconsistent with the input images. In medical settings, such hallucinations pose significant risks, as false pathological findings could lead to unnecessary interventions or missed diagnoses, directly impacting patient care.

In this paper, we introduce ReXTrust, a white-box model designed to detect hallucinations in LVLM-generated radiology reports. ReXTrust uses a self-attention module trained on LVLM hidden states, enabling fine-grained insights into specific radiological findings and reliable overall hallucination risk scores. By analyzing internal model representations, ReXTrust can identify potential hallucinations during the generation process itself, rather than relying solely on post-hoc analysis. We evaluate ReXTrust on a subset of the MIMIC-CXR dataset, demonstrating its effectiveness across different medical categories and severity levels, with particular attention to clinically significant findings that would impact patient care. Our contributions are threefold:

\begin{enumerate}
    \item We develop a white-box architecture for radiology finding hallucination detection that enables both token-level attention analysis and finding-level prediction, making the detection process transparent to clinical users.
    \item We demonstrate state-of-the-art performance in detecting hallucinations in radiology reports, particularly for clinically significant findings, achieving superior results compared to existing approaches while maintaining interpretability.
    \item We provide empirical evidence that model hidden states contain reliable signals for hallucination detection, suggesting a path toward improved semantic fidelity in medical report generation systems.
\end{enumerate}

\section{Background and Related Work}

\subsection{Hallucination Detection}
Hallucination detection is the task of identifying AI-generated content that is false, unfounded, or inconsistent with input data. In medical report generation, hallucinations can manifest as either fictional findings or critical omissions, both of which pose significant risks to patient care. Hallucinations are particularly consequential in radiology, as false findings may both initiate unnecessary medical procedures and mask critical pathologies requiring urgent intervention. Current hallucination detection methods for LLMs and LVLMs can be categorized into three approaches based on their required level of access to model parameters: black-box, gray-box, and white-box approaches. Each category presents different tradeoffs between computational complexity, ease of implementation, and detection accuracy. Our work builds upon these foundations by introducing a white-box approach specifically designed for the radiology domain, where the stakes of hallucination detection are high.

\subsection{Black-Box Methods}
Black-box methods operate without access to model parameters, relying solely on model outputs. Despite this limitation, these methods have gained prominence due to their broad applicability across proprietary models and APIs. Research in this area has pursued two main directions. The first explores models' self-evaluation capabilities, with \citet{kadavath2022language} and \citet{lin2022teaching} demonstrating that LLMs can identify their own hallucinations with reasonable accuracy when explicitly prompted. The second direction quantifies model uncertainty by analyzing output diversity. \citet{kuhn2023semantic} and \citet{farquhar2024detecting} showed that semantic entropy—the uncertainty in the meanings of model-generated text—correlates strongly with hallucination likelihood. Building on these foundations, \citet{friel2023chainpoll} proposed using auxiliary LLMs as external validators, demonstrating improved detection performance across both open and closed-domain tasks. More recently, \citet{manakul2023selfcheckgpt} introduced SelfCheck GPT, which evaluates generated sequences through mutual entailment analysis across multiple generations, providing a more robust assessment of content reliability. While these methods provide valuable insights, our work demonstrates that access to model parameters can facilitate robust hallucination detection in the medical domain.

\subsection{Gray-Box Methods}
Gray-box methods leverage access to token-level probability distributions, enabling more precise analysis than black-box approaches while remaining computationally efficient. Traditional metrics in this category include perplexity, token entropy, and mutual information \citep{fomicheva2020unsupervised, van2022mutual}. However, token-level uncertainty evaluation presents inherent challenges, as large uncertainty values may reflect the presence of multiple viable continuations rather than potential fabrications. To address this limitation, \citet{fadeeva2024fact} developed a claim-conditioned scoring system (CCP) that evaluates model generations according to the probability that a semantically equivalent output would be generated instead, providing a more reliable measure of content veracity. While gray-box methods offer valuable insights into language model uncertainty, our approach moves beyond probability distributions to leverage the rich information contained in hidden states, enabling more nuanced hallucination detection capabilities.

\subsection{White-Box Methods}
White-box methods require complete access to model weights and typically leverage intermediate representations of the model inputs. The predominant approach involves training supervised classifiers on extracted model activations. For instance, \citet{azaria2023internal} and \citet{su2024unsupervised} employed feedforward neural networks for post-generation hallucination detection, while \citet{alnuhait2024factcheckmate} developed similar architectures for pre-generation detection and mitigation. In a significant advancement, \citet{kossen2024semantic} demonstrated that a logistic regression model trained on hidden activations can effectively predict semantic entropy, suggesting that LLMs encode uncertainty information within their internal representations. Building on this insight, \citet{chen2024inside} introduced the EigenScore metric, which analyzes the consistency of hidden state embeddings across multiple generations to provide a robust measure of output reliability. ReXTrust extends previous white-box approaches by using a self-attention mechanism that analyzes sequences of hidden states, thus achieving high fidelity on finding-level hallucination detection.

\subsection{Multimodal Methods}
Given that LVLMs incorporate language model components, hallucination detection methods developed for LLMs can be naturally extended to LVLMs \citep{li2024reference}. However, the visual component of LVLMs presents unique opportunities to validate generated content against input images, particularly for verifying visual claims. Recognizing this potential, \citet{yin2023woodpecker} developed Woodpecker, which employs auxiliary object detectors and visual question-answering models to verify visual assertions in model-generated text. Similarly, \citet{chen2024unified} approached LVLM hallucination detection through independent visual evidence-gathering modules that systematically confirm or contradict claims extracted from model generations. This approach was further refined by \citet{fei2024fine}, who employed Siamese networks to compare scene graphs extracted from model generations with those derived from input images.

In the medical domain, specialized LVLM hallucination detection methods have emerged to address domain-specific challenges. \citet{chen2024detecting} introduced MediHallDetector for categorizing medical hallucinations, demonstrating superior consistency compared to GPT baselines when evaluating medical content. A significant advance was made by \citet{zhang2024radflag} with RadFlag, an approach that leverages entailment relationships between findings generated at different sampling temperatures to identify potentially hallucinatory content in LVLM-generated medical reports. ReXTrust builds upon and complements RadFlag by providing fine-grained insights at both the token and finding levels; our empirical results demonstrate the benefits of analyzing model hidden states over post-hoc output analysis.

\section{Methodology}
In this section, we formalize the task of hallucination detection for radiology report generation. A comparative analysis of black-box, gray-box, and white-box approaches by \citet{liu2024uncertainty} indicated that white-box methods generally demonstrate superior performance. Motivated by this result, we develop ReXTrust as a white-box hallucination detection model.

Throughout our experiments, we utilize the MedVersa model \citep{zhou2024generalist} to generate and evaluate candidate reports. Given a sequence of input chest X-rays, MedVersa generates a candidate report $R$, which we decompose into a set of findings $\{s_i\}_{i = 1}^n$. A finding is defined as a single claim in $R$ (e.g., ``There is pneumonia"), which usually corresponds to a single sentence. In cases where a sentence contains multiple claims, we split the sentence into individual claims. For instance, the sentence ``No evidence of pneumonia, pneumothorax, or pleural effusion" would be split into three distinct findings: ``No evidence of pneumonia," ``No evidence of pneumothorax," and ``No evidence of pleural effusion."

\begin{figure*}[t]
\centering
\includegraphics[width=\textwidth]{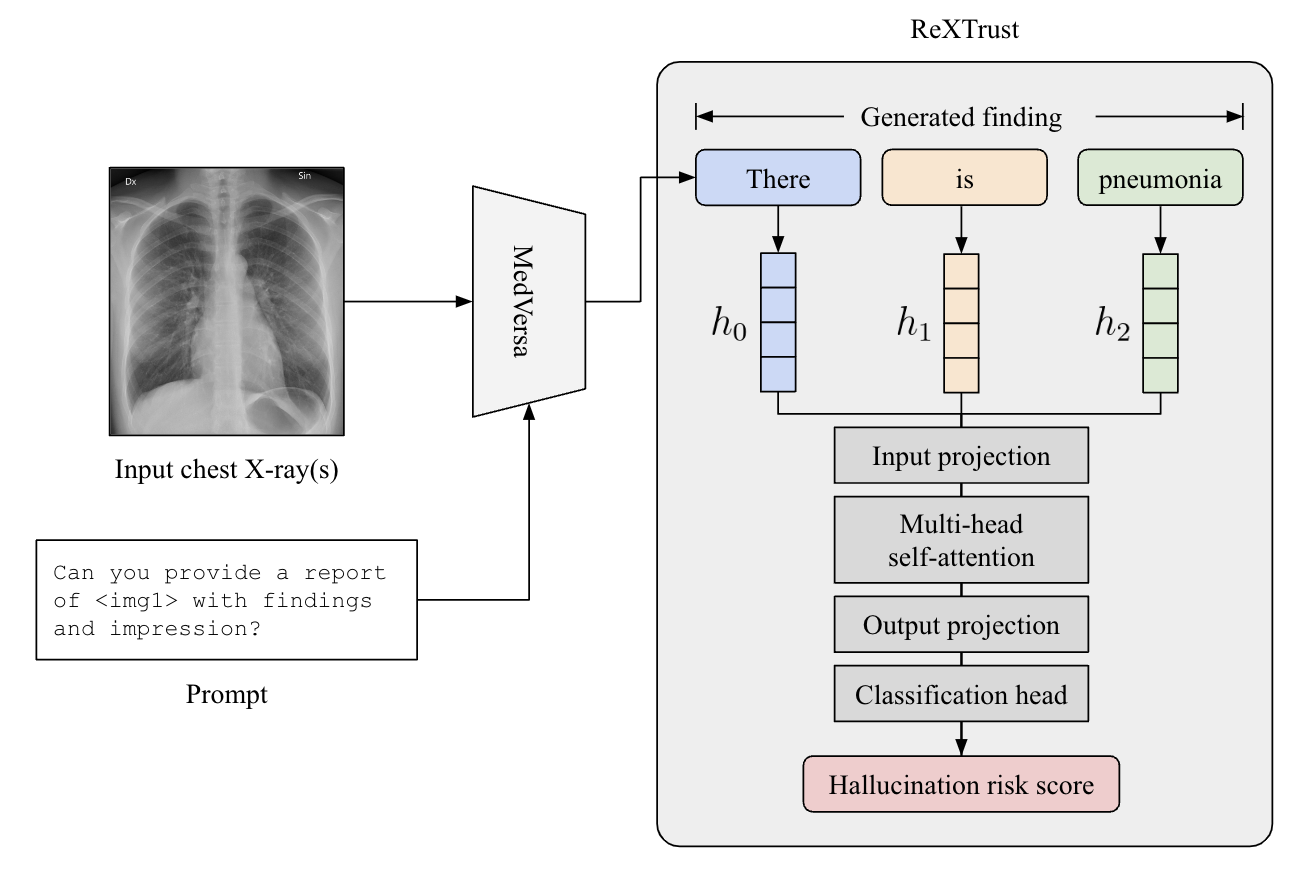}
\caption{ReXTrust hallucination detection framework. The model processes MedVersa's hidden states $h_i$ through a self-attention module to produce finding-level hallucination scores.}
\label{fig:pipeline}
\end{figure*}

\subsection{Model Architecture}

For each individual finding $s_i$, ReXTrust predicts a score in the interval [0, 1], interpreted as the probability that $s_i$ contains hallucinatory content. ReXTrust generates these hallucination risk scores through an end-to-end architecture that processes MedVersa's hidden activation states. The model employs a self-attention module to analyze the sequence of hidden states corresponding to each finding, enabling it to capture both local patterns and broader contextual relationships between tokens when computing the final risk score for $s_i$. Figure \ref{fig:pipeline} provides a schematic representation of this framework. 

For a given finding $s_i$, ReXTrust first extracts the sequence of hidden states $(h_t^l)_{t\in s_i}$ for all tokens $t$ in $s_i$, where $l$ denotes the index of a specific hidden layer (we use $l = 16$ in our implementation) and $h_t^l \in \mathbb{R}^d$ ($d =$ 4096 for MedVersa). This sequence is projected to a 1024-dimensional latent space through a linear transformation and combined with sinusoidal positional embeddings to encode relative token positions. The model then processes this embedding through three 1024-dimensional linear projections to produce query, key, and value vectors. These vectors are fed through an 8-headed self-attention module with dropout regularization ($p = 0.1$), with each head having dimension 128. The attended outputs are mean-pooled along the sequence dimension, passed through another 1024-dimensional projection, and ultimately fed to a sigmoid classification head that produces a hallucination risk score $r$.

The advantages of the ReXTrust architecture are two-fold. First, it enables efficient training and inference while maintaining interpretability for clinical users through attention map analysis. Second, the self-attention mechanism naturally captures the full range of token interactions within a finding, allowing the model to simultaneously consider both local features and long-range dependencies when assessing hallucination risk.

\subsection{Data}
We conduct our experiments using studies from the test set of the MIMIC-CXR database \cite{johnson2019mimic}. To ensure rigorous evaluation, we randomly partition the subjects into a training set for ReXTrust and a held-out set for final evaluation. The training set contains 231 subjects (corresponding to 1923 studies), while the held-out set contains 58 subjects (424 studies). We maintain strict subject-level separation between all sets to prevent data leakage.

For hallucination labeling at the finding level, we adopt the LLM entailment strategy proposed by \citet{zhang2024radflag}. For each finding generated by MedVersa, we employ OpenAI's {\tt gpt-4o} model to classify it as either ``completely entailed," ``partially entailed," or ``not entailed" by the corresponding ground truth radiologist-written report. We consider a finding to be hallucinated if it is either partially entailed or not entailed by the ground truth report.

\subsection{Model Training}
ReXTrust is trained using the average binary cross-entropy between predicted hallucination scores and the associated finding-level hallucination labels as the objective function. We employ a weighted random sampler to ensure balanced exposure to positive and negative examples during training, while maintaining equal class weights in the objective function.

The model is optimized for 5 epochs using the AdamW optimizer \citep{loshchilov2017decoupled} with learning rate $1.0 \times 10^{-4}$ on a cosine schedule, batch size 128, and standard parameters ($\beta_1 = 0.9$, $\beta_2 = 0.999$) with weight decay 0.01. We employ 5-fold cross-validation on the training set, with model performance monitored via the area under the receiver operating characteristic curve (AUROC) on the validation set. The final weights for ReXTrust are obtained by averaging across folds.

\subsection{Finding Severity}
\label{sec:finding_severity}
Radiological findings exhibit substantial heterogeneity in their clinical significance. For instance, a procedural observation such as ``Single portable view of the chest" carries minimal risk if hallucinated, whereas the false positive finding ``There is a left apical pneumothorax" could precipitate unnecessary emergency intervention. To systematically analyze this variance in clinical risk, we employ {\tt gpt-4o} to categorize the findings in our evaluation set into four distinct severity tiers:
\begin{enumerate}
\item Findings requiring immediate emergency intervention.
\item Findings warranting non-emergency clinical action.
\item Findings with minimal clinical significance.
\item Findings which do not fit into the three previous tiers.
\end{enumerate}
The classification is performed using a standardized prompt (detailed in Appendix \ref{app:prompt}). Our analysis encompasses both the complete evaluation set and a focused evaluation of high-risk findings (tiers 1 and 2), enabling a nuanced assessment of hallucination detection performance in clinically critical scenarios. The reliability of these LLM-generated severity labels is discussed in Section \ref{sec:reliability}.

\section{Results}
\subsection{Discriminative Power of ReXTrust}
On the complete set of findings, ReXTrust achieves an AUROC of 0.8751 (95\% CI: 0.8623, 0.8880). When evaluated exclusively on clinically relevant findings, the AUROC improves to 0.8963 (95\% CI: 0.8824, 0.9091). Notably, ReXTrust's performance remains robust—and even shows improvement—when restricted to clinically significant findings. This suggests that ReXTrust learns generalizable features for hallucination detection rather than overfitting to patterns specific to less critical findings.

\subsection{Comparison to Other Approaches}
\label{sec:comparison}
We evaluate ReXTrust against a set of baseline methods spanning diverse approaches to hallucination detection. Traditional uncertainty estimation methods include entropy, which measures token-level predictive uncertainty, and CCP scores \citep{fadeeva2024fact}, which evaluate generations based on the probability semantically equivalent alternatives. Among general-purpose multimodal detectors, we evaluate UNIHD \citep{chen2024unified}, which employs independent evidence-gathering modules to validate generated content. Since UNIHD was not originally designed for the medical domain, we implement domain-specific adaptations that preserve its core mechanisms while enabling fair comparison on radiology data (detailed in Appendix \ref{app:adaptations}). We also compare against EigenScore \citep{chen2024inside}, a white-box method that assesses hallucination likelihood by analyzing similarities between hidden state activations across multiple model-generated sentences (in our case, we use high-temperature samples from MedVersa). Finally, we compare against RadFlag \citep{zhang2024radflag}, a radiology-specific method that identifies hallucinations by analyzing entailment relationships between findings generated at different sampling temperatures.

Table \ref{tab:comparison} presents the comparative evaluation using the micro-averaged AUROC, area under the precision-recall curve (AUPRC), and area under the generalized risk coverage curve (AUGRC) metrics \cite{traub2024overcoming}. ReXTrust demonstrates substantial improvements over all baseline methods across all metrics. The strongest competing method, RadFlag, achieves a finding-level AUROC of 0.7999 (95\% CI: 0.7844, 0.8152), which is 7.52 points lower than ReXTrust. ReXTrust's superior performance relative to UNIHD suggests that training a supervised model directly on in-domain hallucination detection using hidden states provides stronger signals than relying on external verification modules. Furthermore, its improvement over RadFlag indicates that analyzing the generation process through hidden states may be more reliable than post-hoc analysis of model outputs.

To validate our architectural choice of self-attention, we evaluate a variant of ReXTrust that processes hidden states independently for each token rather than attending to the full sequence. In this attention-free version, we generate token-level predictions that are then mean-pooled to produce a finding-level hallucination risk score. While this simpler architecture achieves high performance, it is worse than the full self-attention model. This gap suggests that ReXTrust benefits from modeling the contextual relationships between tokens, capturing dependencies that are difficult to identify through token-level analysis alone.

\begin{table*}[htbp]
\centering
\renewcommand{\arraystretch}{1.2}
\begin{tabular}{lccc}
\toprule
\textbf{Model} & \textbf{AUROC} & \textbf{AUPRC} & \textbf{AUGRC} \\
\midrule
Entropy & 0.7361 (0.7168, 0.7551) & 0.6848 (0.6582, 0.7119) & 0.2093 (0.1987, 0.2202) \\
CCP \citep{fadeeva2024fact} & 0.6535 (0.6334, 0.6734) & 0.5692 (0.5430, 0.5969) & 0.1751 (0.1659, 0.1860) \\
EigenScore \citep{chen2024inside} & 0.6029 (0.5821, 0.6245) & 0.5290 (0.5023, 0.5612) & 0.1979 (0.1878, 0.2086) \\
UniHD \citep{chen2024unified} & 0.6569 (0.6401, 0.6733) & 0.5710 (0.5475, 0.5970) & 0.1638 (0.1528, 0.1719) \\
RadFlag \citep{zhang2024radflag} & 0.7999 (0.7844, 0.8152) & 0.7429 (0.7184, 0.7653) & 0.0988 (0.0907, 0.1067) \\
ReXTrust (without self-attention) & 0.8683 (0.8555, 0.8808) & 0.8186 (0.7958, 0.8431) & 0.0646 (0.0582, 0.0707) \\
\textbf{ReXTrust} & \textbf{0.8751 (0.8623, 0.8880)} & \textbf{0.8310 (0.8097, 0.8529)} & \textbf{0.0637 (0.0571, 0.0701)} \\
\bottomrule
\end{tabular}
\caption{AUROC, AUPRC, and AUGRC performance of ReXTrust on the held-out evaluation set, compared to baseline hallucination detection methods. ReXTrust significantly outperforms other methods across all metrics. Parentheses show the 95\% CIs computed using a 1,000-sample bootstrap.}
\label{tab:comparison}
\end{table*}

\subsection{Performance Across Medical Categories}
\begin{figure}[t]
\centering
\includegraphics[width=\columnwidth]{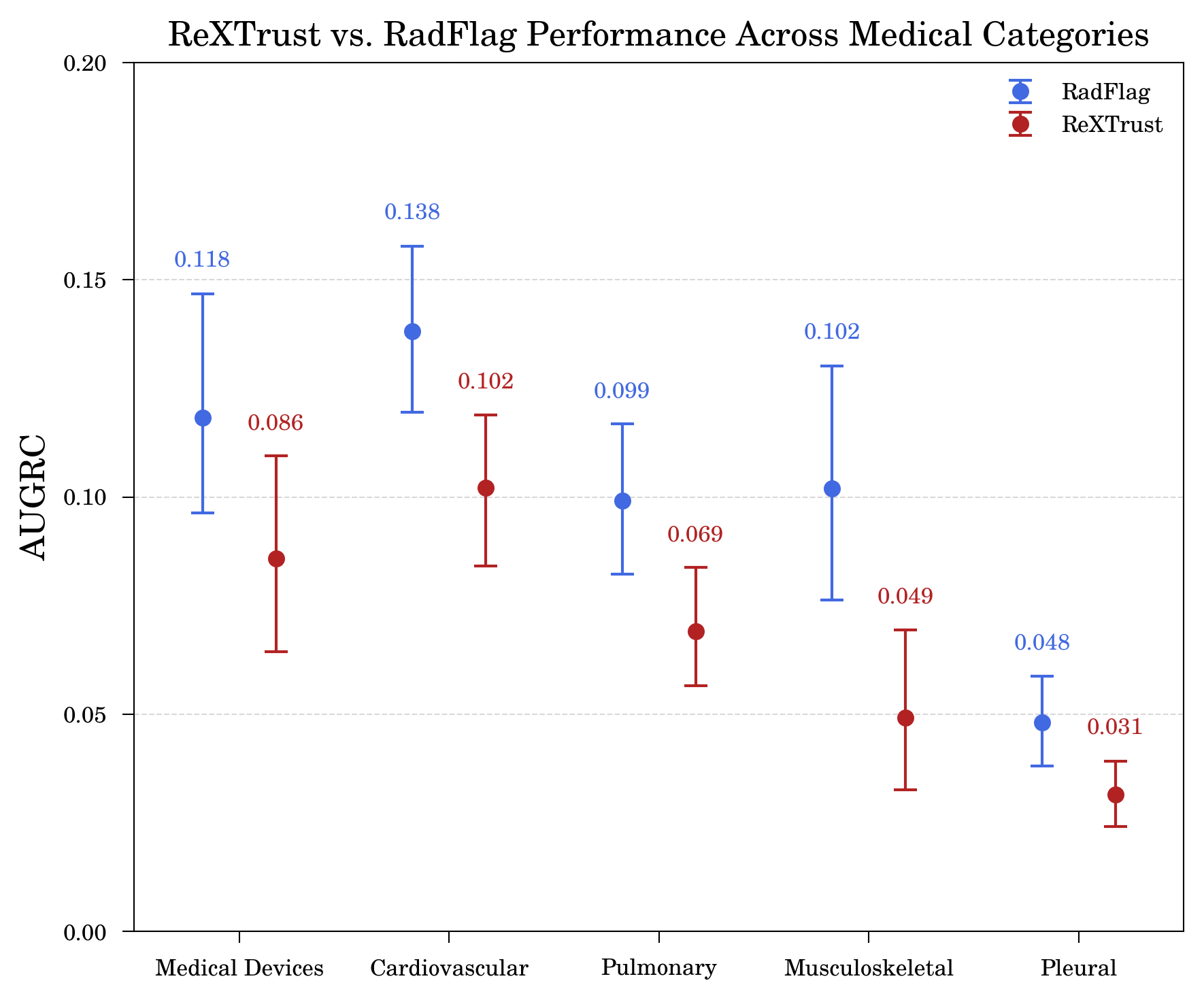}
\caption{ReXTrust AUGRC performance on five medical categories, compared to RadFlag.}
\label{fig:categories}
\end{figure}

Following the categorization approach of \citet{zhang2024radflag}, we refine our analysis by partitioning clinically relevant findings into five distinct categories: Lungs, Pleural, Cardiomediastinal, Musculoskeletal, and Medical Devices. Figure \ref{fig:categories} presents the 95\% confidence intervals for ReXTrust's AUGRC scores (red) on the held-out evaluation set across these categories, compared against those of RadFlag (blue). ReXTrust demonstrates superior selective classification performance across all categories, as indicated by its lower point estimates. To determine whether ReXTrust's performance is statistically significantly better than RadFlag's, we also calculate the confidence intervals on the difference between the AUGRC scores of the two models (Table \ref{tab:diff}). We find that none of the confidence intervals contain zero, and therefore conclude that the selective classification performance of ReXTrust is statistically significantly better than that of RadFlag at the finding level across all categories.

\begin{table}[htbp]
\centering
\renewcommand{\arraystretch}{1.2}
\begin{tabular}{lccccc}
\toprule
\textbf{Category} & $\textbf{AUGRC}_{\text{RadFlag}} - \textbf{AUGRC}_{\text{ReXTrust}}$ \\
\midrule
Medical Devices & 0.0324 (0.0050, 0.0621) \\
Cardiovascular & 0.0359 (0.0151, 0.0551) \\
Pulmonary & 0.0300 (0.0140, 0.0486) \\
Musculoskeletal & 0.0528 (0.0311, 0.0803) \\
Pleural & 0.0166 (0.0082, 0.0253) \\
\bottomrule
\end{tabular}
\caption{Difference in AUGRC between RadFlag and ReXTrust on five medical categories. ReXTrust demonstrates statistically significantly better selective classification performance across all categories.}
\label{tab:diff}
\end{table}

Table \ref{tab:keywords} evaluates ReXTrust on findings containing commonly occurring keywords. For each keyword, we show ReXTrust's micro-averaged $F_1$ score on clinically relevant findings that contain the keyword and on which the model displays high confidence (i.e., the predicted score is either in the top or bottom score quartile for that category). ReXTrust achieves impressive performance on pleural effusion, pneumonia, and edema findings. However, it performs worse on findings related to pneumothorax, atelectasis, and tubes. We also note that ReXTrust performs reasonably well when detecting hallucinations in findings containing positional keywords despite not containing explicit visual modules.

\begin{table}[htbp]
\centering
\renewcommand{\arraystretch}{1.2}
\begin{tabular}{lccccc}
\toprule
\textbf{Keyword} & \textbf{$F_1$} \\
\midrule
``pleural effusion" & 0.8403 (0.7847, 0.8958) \\
``pneumothorax" & 0.6193 (0.5455, 0.6932) \\
``consolidation" & 0.7000 (0.5857, 0.8000) \\
``pneumonia" & 0.9091 (0.8182, 0.9773) \\
``edema" & 0.8857 (0.8000, 0.9571) \\
``atelectasis" & 0.6000 (0.4500, 0.7500) \\
``tube" & 0.5750 (0.4000, 0.7250) \\
``right" & 0.6981 (0.6038, 0.7830) \\
``left" & 0.6591 (0.5679, 0.7614) \\
\bottomrule
\end{tabular}
\caption{Micro-averaged $F_1$ scores of ReXTrust on selected finding keywords.}
\label{tab:keywords}
\end{table}

\section{Discussion}

\subsection{Qualitative Examples}
\begin{figure*}[t]
\centering
\includegraphics[width=\textwidth]{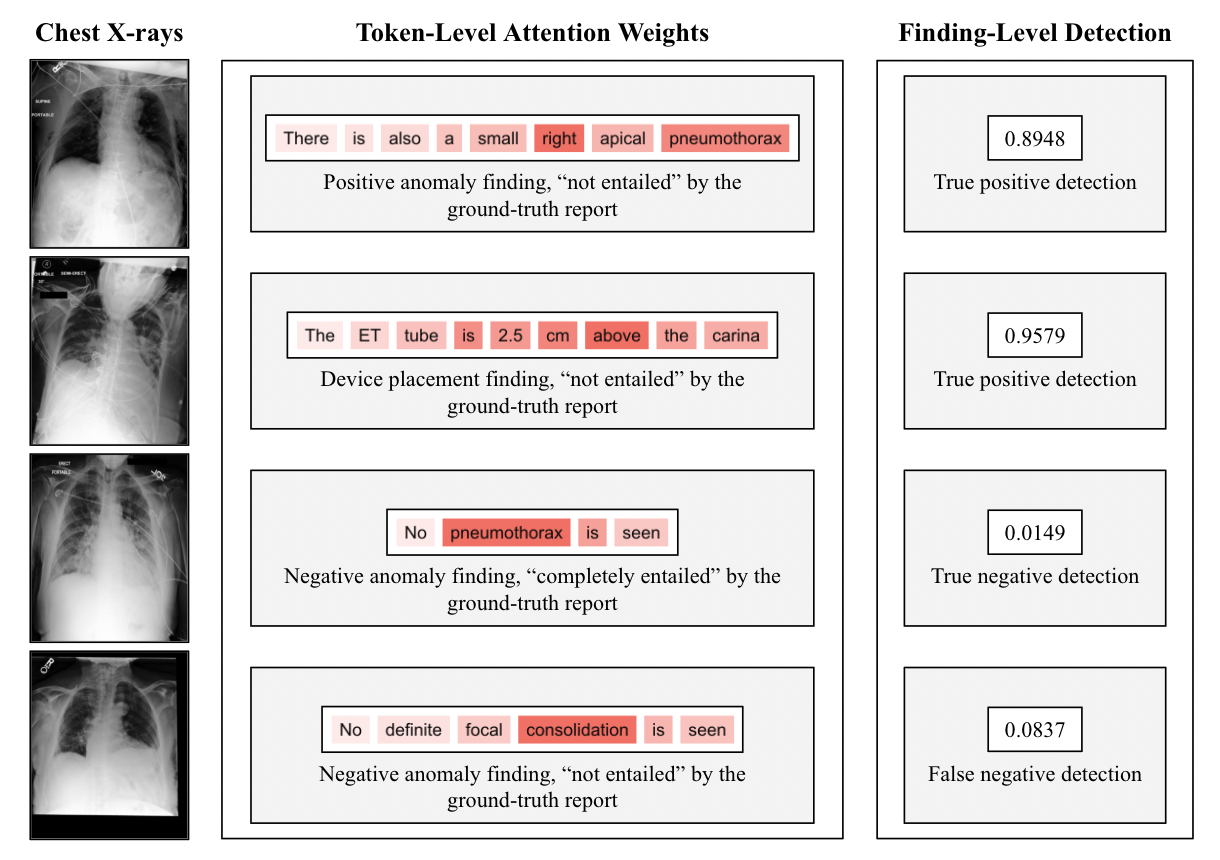}
\caption{Qualitative examples of ReXTrust on four studies from our held-out evaluation set. Shown on the left are examples of chest X-rays serving as inputs to MedVersa. Shown in the middle are findings generated by MedVersa, as well as token-level attention visualizations from ReXTrust. Shown on the right are the finding-level risk scores output by the classification head of ReXTrust.}
\label{fig:qualitative}
\end{figure*}

Figure \ref{fig:qualitative} presents an analysis of ReXTrust's performance on four representative studies from the held-out evaluation set. ReXTrust's self-attention mechanism enables identification of specific semantic components that contribute to the overall hallucination risk assessment. In the first case, ReXTrust assigns high attention weights to ``right" and ``pneumothorax," correctly identifying this positive finding as hallucinatory. 

High attention weights are not necessarily indicative of hallucination risk, however. For instance, the third model-generated finding indicates the lack of an anomaly; ReXTrust's attention weights are highest on the ``pneumothorax" tokens, but its final prediction indicates that the finding is true. In the fourth finding, it is the converse; ReXTrust focuses on the ``consolidation" tokens, but the model falsely predicts that the finding is not hallucinatory.

\subsection{Complementarity with RadFlag}
\label{sec:overlap}
To investigate whether ReXTrust and RadFlag capture complementary aspects of hallucination detection, we evaluate a linear ensemble that combines their predictions with weights 0.8 and 0.2, respectively. The ensemble achieves an AUROC of 0.8822 (95\% CI: 0.8699, 0.8947), an AUPRC of 0.8471 (95\% CI: 0.8276, 0.8670), and an AUGRC of 0.0594 (95\% CI: 0.0532, 0.0656), surpassing the performance of both individual models.

The complementary nature of these approaches stems from their fundamentally different detection strategies: ReXTrust analyzes model hidden states to identify potential hallucinations during the generation process, while RadFlag employs temperature-based sampling to detect inconsistencies in model outputs. Furthermore, the success of the ensemble suggests that ReXTrust could be extended to incorporate hidden states from outputs sampled at high temperatures.

\subsection{Generalizability Across Architectures}
While ReXTrust was trained to detect hallucinations in MedVersa-generated reports using MedVersa hidden states, this framework can be readily adapted to other medical LLMs and LVLMs, such as RaDialog \citep{pellegrini2023radialog} and LLaVA-Med \citep{li2024llava}, and Maira-2 \citep{bannur2024maira}. Although the architecture of ReXTrust would need to be modified to accommodate different hidden state dimensions, the core principle of applying a self-attention module over sequences of hidden states is applicable. As such, models like ReXTrust could serve as valuable tools for assessing the factual reliability of large medical language models. If hallucinations can be reliably predicted from a model's hidden states, it suggests that modifications to the training methodology may be necessary to reduce the frequency of hallucinatory content.

\subsection{Label Reliability}
\label{sec:reliability}
Our study relies on LLM-generated hallucination and severity labels, which inherently contain some degree of noise. Previous work by \citet{zhang2024radflag} evaluated similar hallucination labels through clinical validation, finding high but imperfect agreement between LLM and clinician assessments. We perform a similar evaluation of the severity labels in collaboration with an expert clinician. From a sample of 60 findings across 39 studies—with 15 findings from each of the four severity categories defined in Section \ref{sec:finding_severity}—we find that 8 of the 30 findings categorized into tiers 3 and 4 by the LLM were classified as belonging to tiers 1 and 2 by the clinician. Conversely, none of the 30 findings categorized into tiers 1 and 2 by the LLM were reclassified into tiers 3 and 4 by the clinician. These results suggest that we likely underestimate the number of clinically significant findings. However, given ReXTrust's consistent performance across finding categories, we expect its performance on the true set of clinically significant findings to align with the global results presented in Table \ref{tab:comparison}.

\subsection{Limitations and Future Work}
\label{limitations}
We identify three primary limitations of this study. First, ReXTrust depends on supervised labels. Our work leverages expert-written radiology reports from MIMIC-CXR to generate binary hallucination labels for training. In scenarios where high-quality ground truth reports are unavailable, ReXTrust's white-box approach may be less suitable than unsupervised alternatives. However, given that ReXTrust achieves high performance even when trained on a small dataset, it may still be useful in label-limited settings.

Second, ReXTrust shows suboptimal performance on certain types of radiological findings (see Table \ref{tab:keywords}). While this weakness might be partially mitigated through ensembling with orthogonal approaches, as discussed in Section \ref{sec:overlap}, future work should incorporate more explicit visual grounding tools to verify the factuality of assertions in AI-generated findings \citep{he2024parameter, zou2024medrg, shaaban2024medpromptx, bannur2024maira}.

Third, we note that several benchmarks used for comparison in Section \ref{sec:comparison} were not originally designed for radiology (e.g., UNIHD). Although we implemented domain-appropriate modifications to ensure fair comparison with our approach, we may underestimate their potential performance in the medical domain. Future work should explore extensive adaptations of general-purpose hallucination detectors to medical imaging tasks.

\section{Conclusion}
We have presented ReXTrust, a white-box framework for detecting hallucinations in AI-generated radiology reports. Our approach demonstrates superior performance compared to existing hallucination detection methods, achieving an AUROC of 0.8751 across all findings and 0.8963 on clinically significant findings. ReXTrust's end-to-end architecture enables both granular analysis of hallucination patterns and reliable overall risk assessment.

Our results indicate that model hidden states contain valuable signals about the reliability of generated content that can be effectively leveraged for hallucination detection. The robust performance of ReXTrust on clinically significant findings is particularly noteworthy, as it demonstrates the model's ability to identify potentially harmful hallucinations that could impact patient care. Our work also identifies important areas for future research, such as improving hallucination detection performance on specific classes of findings and developing methods to reduce dependence on supervised labels. Nevertheless, ReXTrust represents a significant advance in ensuring the reliability of medical report generation models. As medical LVLMs continue to evolve and see increased clinical adoption, approaches like ReXTrust will be essential for maintaining high standards of accuracy and safety in medical AI systems.

\appendix
\section{Finding Severity Prompt}
\label{app:prompt}
We present the complete prompt template used to categorize AI-generated findings into severity tiers in Figure \ref{fig:prompt}.

\begin{figure*}
\begin{tcolorbox}[
    colback=lightgray,
    colframe=black,
    boxrule=1pt,
    arc=3pt,
    left=10pt,
    right=10pt,
    top=10pt,
    bottom=10pt,
    width=\textwidth
]
You are an AI radiology assistant helping process reports from chest X-rays. You are given a finding (F) from a chest X-ray report. Your job is to judge and categorize the clinical severity of the finding. For positive findings, here are the definitions for the severity categories you should use:

\begin{enumerate}
    \item Emergency clinical consequence
    \begin{itemize}
        \item Definition: Findings which could lead to outcomes requiring immediate or urgent medical intervention to prevent significant harm, deterioration, or life-threatening complications. This includes scenarios where rapid diagnosis and management are critical to stabilize the patient but excludes conditions that can be managed electively.
        \item Examples: Tension pneumothorax, acute myocardial infarction, Endotracheal tube misplacement compromising airway function.
    \end{itemize}
    \item Non-emergency but actionable clinical consequence
    \begin{itemize}
        \item Definition: Findings which require medical attention or follow-up but do not necessitate immediate or urgent intervention to prevent significant harm. These findings are important for patient health management and should be addressed through routine clinical care, elective procedures, or scheduled follow-ups.
        \item Examples: Newly diagnosed malignancy without acute complications, chronic infection like tuberculosis, significant but non-acute organ abnormalities such as a thyroid nodule needing further investigation.
    \end{itemize}
    \item Clinically insignificant consequence
    \begin{itemize}
        \item Definition: Findings that pose no health risk or adverse outcome for the patient and require no clinical intervention or monitoring. These findings are typically benign, normal variations, or imaging artifacts that do not impact patient management or treatment plans.
        \item Examples: Normal anatomical variations (e.g., azygos fissure), incidental imaging artifacts, findings that are clinically irrelevant to patient health and do not necessitate any follow-up.
    \end{itemize}
    \item Other
    \begin{itemize}
        \item Definition: Any findings that do not fit into the above categories.
    \end{itemize}
\end{enumerate}

For negative findings, you should evaluate the severity of the negations of those findings. For example, if you are given the finding, ``There is no pneumothorax," you should evaluate the severity of the negation, ``There is pneumothorax," and return the corresponding severity category (which would be ``emergency clinical consequence" in this case).
    
Return a JSON of the format: {\tt \{"severity\_category": <category>, "reason": <reason>\}}, where {\tt <category>} is one of ``emergency clinical consequence", ``non-emergency but actionable clinical consequence", ``clinically insignificant consequence", or ``other", and {\tt <reason>} is a string explaining your reasoning for the severity category.
\end{tcolorbox}
\caption{Prompt template for radiological finding severity classification. The prompt provides detailed definitions and examples for each severity category to ensure consistent classification across all experiments.}
\label{fig:prompt}
\end{figure*}

\section{Adaptating UNIHD to Radiology}
\label{app:adaptations}
Noting that UNIHD is not adapted to radiology images and reports, we apply sensible adaptations in order to provide a fair comparison with ReXTrust. Specifically, we modify the prompts of the individual modules to include radiological terminology and examples. We also replace the authors' Grounding Dino object detection module \citep{liu2023grounding} with a custom module based on pre-trained models from \citet{Cohen2022xrv}.

\section{Ablation Studies}

\subsection{Hidden Layer Index}
\begin{figure}[t]
\centering
\includegraphics[width=\columnwidth]{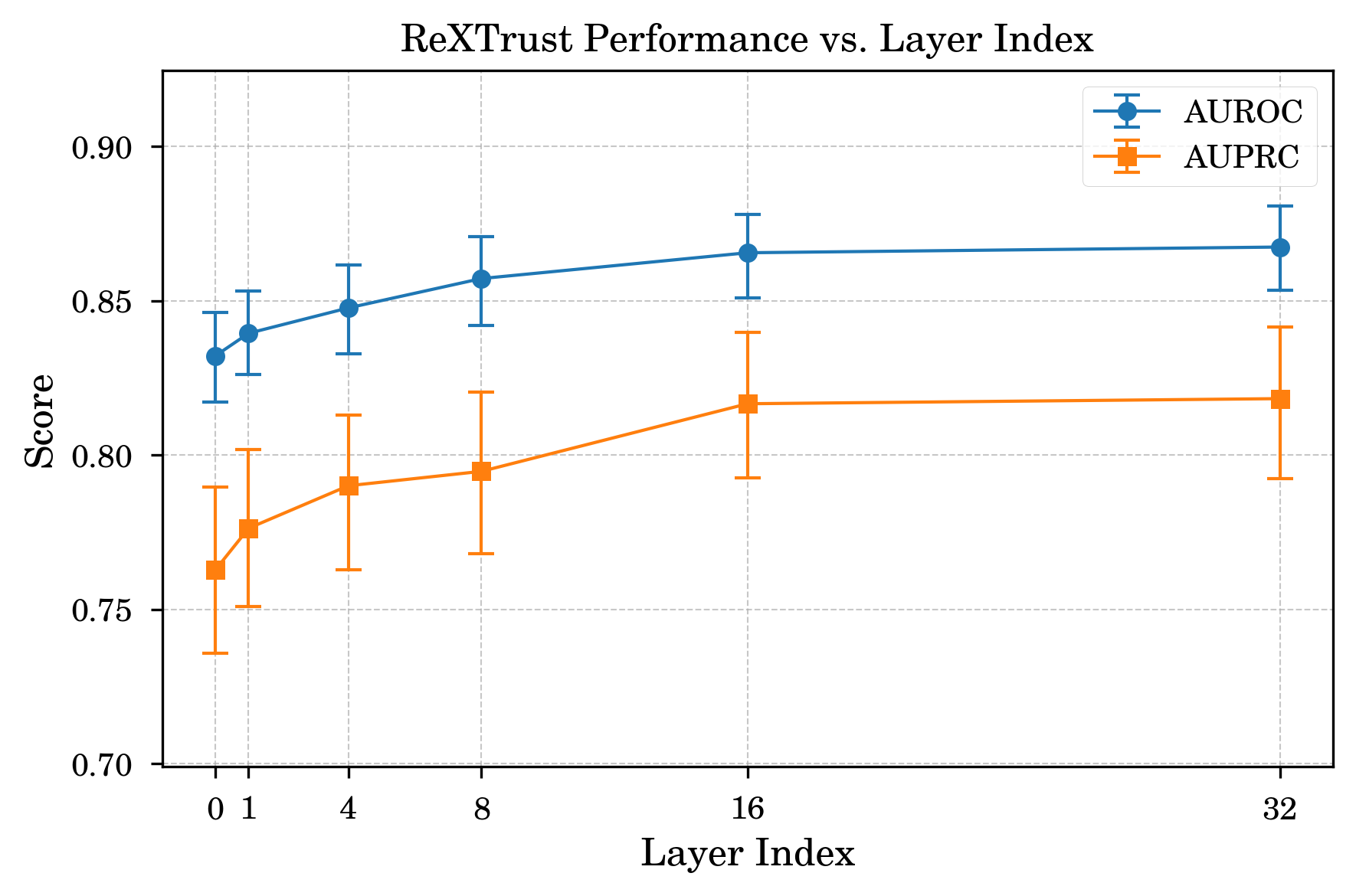}
\caption{AUROC and AUPRC performance of ReXTrust as a function of the layer index of the MedVersa hidden states used as inputs. Model performance saturates near layer 16. Error bars indicate 95\% confidence intervals.}
\label{fig:layer_performance}
\end{figure}

We test the performance of ReXTrust as a function of the hidden layer index of MedVersa. Figure \ref{fig:layer_performance} shows the AUROC (blue) and AUPRC (orange) scores of ReXTrust trained on hidden states $h_t^l$ as we let $l$ vary from 0 to 32. We find that the performance of ReXTrust saturates near layer 16.



\bibliography{bibliography}

\end{document}